%% file: main_cameraready.tex
\documentclass{article}

\usepackage{microtype}
\usepackage{graphicx}
\usepackage{subcaption}
\usepackage{booktabs} 

\usepackage{hyperref}

\usepackage[accepted]{icml2026}

\usepackage{amsmath}
\usepackage{amssymb}
\usepackage{mathtools}
\usepackage{amsthm}
\usepackage[toc,page,header]{appendix}
\usepackage{titletoc}
\usepackage{siunitx}
\usepackage{multicol, multirow}
\usepackage{enumitem}
\usepackage{xspace}
\usepackage{lipsum}

\usepackage[dvipsnames]{xcolor}
\usepackage{tikz}

\usepackage[capitalize,noabbrev]{cleveref}

\theoremstyle{plain}

\theoremstyle{definition}

\theoremstyle{remark}

\input{math_commands}
\newcommand{\methodname}{\texttt{\textbf{ARM}}\xspace}

\usepackage[textsize=tiny]{todonotes}

\icmltitlerunning{Attention with Routed-Memory for Learnable Sparse Control}

\begin{document}

\twocolumn[
    \icmltitle{Attention with Routed-Memory for Learnable Sparse Control}
    
    
    \icmlsetsymbol{equal}{*}
    
    \begin{icmlauthorlist}
    \icmlauthor{Qiuhao Zeng}{equal,uoft,vector}
    \icmlauthor{Jerry Huang}{equal,riken,udem,mila}\\
    \icmlauthor{Peng Lu}{udem}
    \icmlauthor{Ruiyi Fang}{uwo}
    \icmlauthor{Gezheng Xu}{uwo}
    \icmlauthor{Zihao Jing}{uwo}
    \icmlauthor{Yufei Cui}{}
    \icmlauthor{Charles Ling}{uwo}
    \icmlauthor{Gang Niu}{riken}
    \icmlauthor{Boyu Wang}{uwo,vector}
    \end{icmlauthorlist}
    
    \icmlaffiliation{uoft}{University of Toronto, Canada}
    \icmlaffiliation{vector}{Vector Institute, Canada}
    \icmlaffiliation{uwo}{Western University, Canada}
    \icmlaffiliation{udem}{Universit\'{e} de Montr\'{e}al, Canada}
    \icmlaffiliation{mila}{Mila - Quebec AI Institute, Canada}
    \icmlaffiliation{riken}{RIKEN AIP, Japan}
    
    \icmlcorrespondingauthor{Boyu Wang}{bwang@csd.uwo.ca}
    
    \icmlkeywords{KV Cache, Sparse Attention, Differentiable Memory, Gumbel-Softmax, MDP}
    
    \vskip 0.3in
]



\printAffiliationsAndNotice{\icmlEqualContribution}

\begin{abstract}
Despite advances in long-context inference, large language models (LLMs) remain fundamentally limited by the key-value (KV) caching mechanisms that are necessary for stable computation. Techniques such as selective token eviction and pruning have vastly mitigated these issues, but often discard core information to manage the growing cache.
In this paper, we propose \textit{\textbf{A}ttention with \textbf{R}outed \textbf{M}emory} (\texttt{\textbf{ARM}}) a novel KV caching structure that introduces a fully differentiable, fixed-size memory system organized as a hierarchical router. Via a Gumbel-Softmax, \texttt{\textbf{ARM}} learns to select memory slots and perform sigmoid-gated updates that softly combine new and stored information, avoiding hard eviction and reducing information loss. By further training a policy to dynamically select varying amounts of memory at inference, \texttt{\textbf{ARM}} adapts its accesses for both simple contexts and inputs that require deeper reasoning, enabling more scalable and effective retrieval on both short- and long-contexts.
    Experimental results on standard commonsense and long-context reasoning benchmarks demonstrate that \texttt{\textbf{ARM}} achieves superior performance and efficiency compared to fixed KV-caching approaches, while remaining efficient and scalable in terms of both memory and generation latency.
\end{abstract}

\section{Introduction}

Being able to handle longer context has become a large focus within the realm of context maintenance and attention mechanism scalability, with many \textit{large language models}~(LLMs) now demonstrating an ability to handle such settings~\citep{gpt4, claude3, gemini2.5, qwen3, command-a}. However, challenges remain. In particular, attention computation remains hindered by the growing size of the cache due to attention calculation across past KVs, leading to growing decoding latency and memory usage that prohibit long-context inference~\citep{longbench,attention-sinks}. As context lengths extend, the memory requirements for storing these cached states grow linearly, rapidly surpassing the memory footprint of the model itself. This ballooning memory consumption not only escalates deployment costs by necessitating high-memory hardware but also severely degrades inference latency~\citep{kvquant,shutova2025cache}. Because the sheer volume of cached data saturates memory bandwidth, the generation process transitions from being compute-bound to memory-bound.

Addressing this concern has led to a number of approaches that can be categorized into three fundamental methodologies: (1) {\textit{quantization}} techniques~\citep{kivi, kvquant, zeroquant, kv1bpc, specache}, which reduce numerical precision while preserving semantic integrity, allowing for the storage of larger caches; (2) {\textit{pruning}} mechanisms~\citep{h2o, attention-sinks, snapkv, flexgen}, which selectively retain critical tokens to maintain a bounded cache size; and (3) {\textit{offloading}} strategies~\citep{infinigen, quest, loki, palu, ld-projected-attention}, which redistribute memory across heterogeneous storage hierarchies. Hybrid methods~\citep{shadowkv} further combine multiple techniques to maximize efficiency. These have made profound improvements in inference efficiency, but each maintains a fundamental limitation, namely a still-increasing cache, the risk of of discarding important information, and latency resulting from inter-hierarchy transfer.

To bridge this gap, we introduce \textit{Attention with Routed Memory} ({\methodname}), a pruning-inspired approach that uses a memory structured as a novel hierarchical router to manage the KV cache during inference. Unlike fixed-rule token eviction policies, \methodname employs a structured routing mechanism with a fixed set of writable memory slots where token-level information is \textit{softly compressed} via learnable updates rather than explicitly discarded. This design ensures a constant cache size while guaranteeing the accessibility of historical information. This is further augmented with a dynamic selection policy that, rather than retrieving the full KV cache, learns to adapt the number of memory buckets retrieved on a per-input basis and per-layer basis, enabling sparse access for simple queries and expanded retrieval when more complex or information-intensive reasoning is required. Consequently, \methodname unifies efficient sparse attention control with dynamic memory networks~\citep{neural-turing-machine, memory-networks, chandar2016hierarchical}. More concretely, our contributions are:
\begin{enumerate}[leftmargin=10pt,label=\roman*),topsep=0pt,itemsep=0pt,parsep=2pt]
    \item \textbf{Learnable Soft Eviction:} 
    We propose a learnable eviction policy for KV-cache management that uses Gumbel-Softmax~\citep{gumbel-softmax} routing to select memory locations and update these locations with sigmoid gating to softly integrate new information without hard eviction, enabling better memory preservation in a bounded cache.
    \item \textbf{Adaptive Top-$M$ Retrieval:} 
    We introduce a learnable retrieval policy that dynamically selects the number of retrieved memory buckets based on the input context, allowing attention sparsity to adapt to task complexity rather than relying on fixed or hand-crafted retrieval heuristics.
    \item \textbf{Scalable Long-Context Inference:} We motivate why shared writes alongside learnable sparsity control enable more scalable long-context inference compared to existing approaches, proved through experimental findings.
\end{enumerate}

Our results on a series of benchmarks~\citep{eval-harness, longbench, ruler} demonstrate the effectiveness of this approach, highlighting the potential benefits of such alternative key-value caching structures, with further benchmarking showing additional advantages in memory usage and latency. In the era of growing LLM complexity and usage, the marked improvements from our method suggest a pathway towards greater efficiency and scalability through better motivated structural design.

\begin{figure*}[t!]
    \centering
    \vspace{-0.5\baselineskip}
    \includegraphics[width=\linewidth]{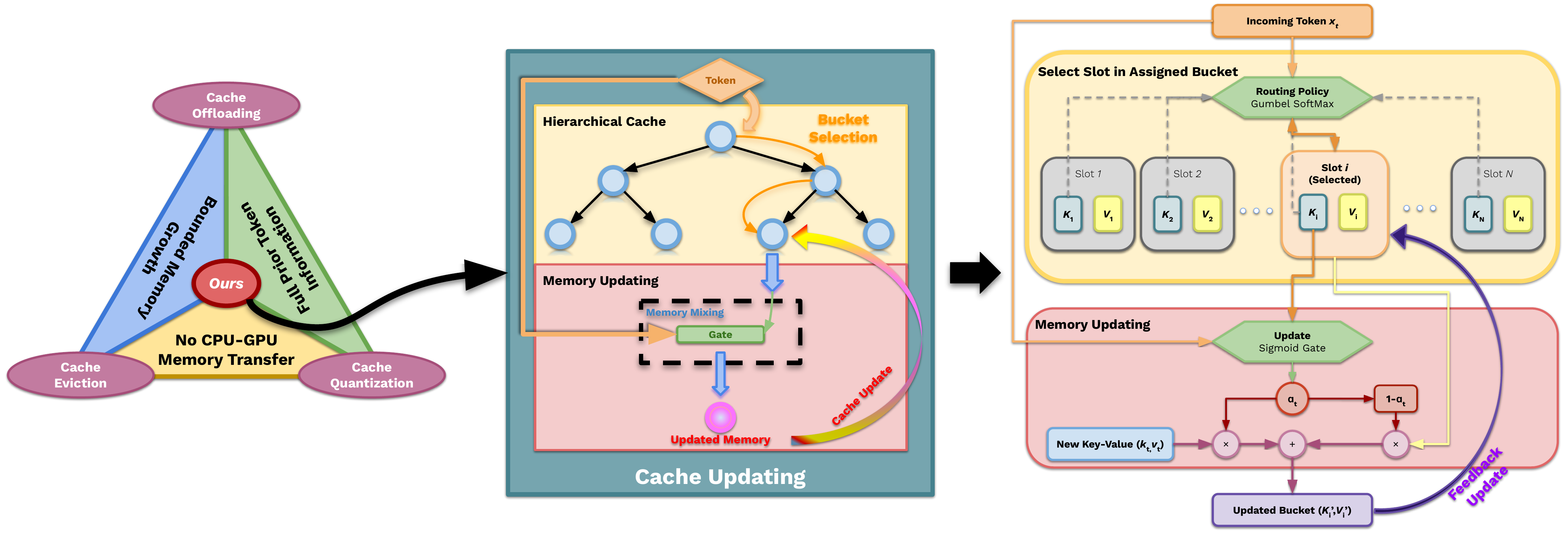}
    \vspace{-1.5\baselineskip}
    \caption{Outline of our \methodname caching structure and the space it falls within the different caching paradigms. For any incoming token, memory slots are selected from the hierarchical cache using a memory controller. The selected slots are used for attention computation. At the same time, the memory slot is updated through a gated memory update.}
    \vspace{-\baselineskip}
    \label{fig:method}
\end{figure*}

\section{Related Work}

\subsection{KV-Cache Management and Token Eviction}

KV caches prevent the re-computation of keys and values by storing them in memory during model inference. Managing this memory footprint becomes imperative, as its unbounded growth becomes a bottleneck within attention computation. This has led to two primary management strategies to handle the footprint. The first is to store the KV cache in a quantized state; this lower precision enables for a greater raw number of KV pairs to be stored within the same memory capacity, effectively increasing practical context capacity~\citep{kivi, kvquant}. An alternative is to remove the concern of a fixed memory capacity by instead bounding the number of pairs that can be stored through the use of eviction policies that eliminate KV pairs once the maximum size is reached, done either through heuristic approaches~\citep{h2o, attention-sinks} or more dynamic selection~\citep{d2o, dmc, less,qin2025cake}. However, these existing eviction strategies universally trade off either long-range context fidelity or computational simplicity by relying on rigid windows~\citep{attention-sinks}, coarse heuristics~\citep{adakv}, or expensive per-head clustering~\citep{snapkv}, therefore resulting in non-negligible approximation error or management overhead. Alternatively, our approach strikes a balance by using a fixed size structure that dynamically mixes incoming KV pairs.

\subsection{Sparse Attention and Dynamic KV Selection}

To overcome the quadratic complexity of attention~\citep{attention}, the use of location-based sparse patterns to compute sparse attention has been widely adopted as a manner of reducing this complexity~\citep{top-k-attention, beltagy2020longformer, mao2024iceformer, xiong2021nystromformer, zeta, zaheer2020big}. This has led to similar approaches in KV-cache management, primarily through the use of dynamic sparse KV selection. In these settings, attention is selectively computed on a subset of tokens within the KV cache while the cache itself is maintained in its entirety. Diverse token selection approaches exist with varying degrees of efficacy; however, a primary feature of such methods is the need to offload the cache to CPU memory~\citep{infinigen, shadowkv, lrqk}, which offers larger overall storage space, while loading only the relevant pairs back to GPU memory for attention computation, leading to greater overall latency due to bottlenecks in transfer bandwidth. In contrast, our approach maintains information from all past key-value pairs within the fixed set of memory slots, enabling for full storage within GPU memory, while also offering sparse attention computation through its routing-based tree structure.

\subsection{Attention as Neural Memory}
Given a sequence of $T$ tokens $\mX\!=\!\left[\vx_1,\dots, \vx_T\right]^\top\!\in\!\mathbb{R}^{T \times d}$, causal self-attention is computed as:
\begin{equation}
  \label{eq:standard-attention}
  \mA = \operatorname{softmax}\left((\mQ \mK^\top)\odot\mathbf{M}\right)\mV,
\end{equation}
where $\mQ,\mK,\mV\!\in\!\mathbb{R}^{T\times d}$ are mappings of $\mX$ via weights $\mW_q,\mW_k,\mW_v\!\in\!\mathbb{R}^{d\times d}$,
$\mathbf{M}\!=\!\left\{M_{ij}\!=\!1 \operatorname{if} i \geq j \operatorname{else}-\infty\right\}$ is the mask to prevent future information leakage and $\odot$ denotes element-wise multiplication.

At any given timestep $t$, $\mK,\mV$ can be viewed as neural \emph{key-value memories}~\citep{memory-networks, geva-etal-2021-transformer} $\widetilde{\mK}_t,\widetilde{\mV}_t\!\in\!\mathbb{R}^{m\times d}$, with $m$ memory slots. At step $t$, the query $\vq_t\!=\!\mW_q\,\vx_t\!\in\!\mathbb{R}^{d}$ first attends to the key memories $\widetilde{\mK}_t$ to retrieve relevant information, which is then summarized into $\vo_t$ by computing a weighted sum of the value memories $\widetilde{\mV}_t$ \citep{zhang-cai-2022-linearizing} using the normalized attention scores:
\begin{equation} \label{eq:memory}
    \vo_t = {\widetilde{\mV}_t}^\top \operatorname{softmax}({\widetilde{\mK}_t} \vq_t).
\end{equation}
From this perspective, Transformers are equipped with an unbounded number of memory slots, which grow linearly w.r.t the sequence length \citep{oren-2024-transformers} (i.e., $m\!=\!t$ for step $t$) -- a new key $\vk_t\!=\!\mW_k\,\vx_t\!\in\!\mathbb{R}^{d}$ is assigned with a unique memory slot upon its introduction, leading to a memory update rule: $\widetilde{\mK}_t\!=\!\widetilde{\mK}_t\!\cup\!\{\vk_t\}$, with $\widetilde{\mV}_t$ updated similarly.
This, however, comes at the cost of quadratic time complexity w.r.t. $T$ for training and $O(T\,d)$ time/memory complexity at inference \citep{pope-2022-efficiently}, posing challenges for large-scale models.

From this perspective, fixing the number of memory slots to a constant size \(m\!\ll\!T\) can reduce complexity. \citet{abc} propose the Attention-with-Bounded-Memory-Control (ABC) mechanism to allow writing multiple tokens into each memory slot, while \citet{gsa} adapts this for linear models by incorporating a gating mechanism inspired by \citet{gla}. Nevertheless, these do not account for more fine-grained memory control at inference, leading to potential issues from disproportionate information mixing. In contrast, our approach combines a fixed-size memory along with information theoretic gains from sparse attention to resolve this.

\section{Methodology}

We introduce Attention with Routed-Memory, \methodname, which we intend as a balance between current KV-cache management mechanisms. \methodname makes use of a hierarchical memory structure, where leaf nodes represent individual memory slots in which information is stored. When a new query token arrives, the token is routed through the memory structure to determine the appropriate memory slots to compute attention with, while also selecting a new memory slot in which information about this query will be mixed into softly. The following sections define the main components of \methodname.

\subsection{Hierarchical Routed Memory}

We organize the KV cache as a \emph{hierarchical routed memory} $\mathcal{M}$, instantiated as a fixed multi-level tree that directly parameterizes a persistent, bounded cache. The tree consists of internal routing nodes and $N_b$ leaf nodes, referred to as \emph{buckets}. Each node in the hierarchy functions as a router that computes routing logits over its outgoing branches, analogous to routing mechanisms used in mixture-of-experts models \citep{shazeer2017outrageously, lepikhin2020gshard}.

Given an incoming token, routing proceeds deterministically from the root to a leaf bucket by selecting, at each level, the child node with the largest routing logit. This top-down routing assigns each token to a unique leaf bucket. Each bucket contains a fixed number of KV slots that store representations of tokens routed to that leaf. The router functions are trained using a self-supervised learning objective~\citep{anonymous2026hierarchical}, encouraging semantically similar tokens to be assigned to the same bucket. The objective is defined as a two-part balancing loss to (1) each token's assignment probability at any given level is approximately similar and (2) enforce that the number of tokens routed to each node is similar. More explicitly, 
\begin{align*}
\mathcal{L}_1
= \frac{1}{T} \sum_{i=1}^{T} H(\vp_{i,p}^{(l)}), &\quad \mathcal L_2
= \sum_{p} \sum_{j=1}^{C} \bar p_{p,j}^{(l)} \log \bar p_{p,j}^{(l)} \\
\mathcal{L}_{\mathrm{bal}} &= \mathcal{L}_1 + \mathcal{L}_2
\end{align*}
where $\vp_{i,p}^{(l)}$ is the probability vector of assignments of token $i$ at the $(l)$-th level of the router and $\bar p_{p,j}^{(l)}$ is the assignment probabilities of tokens in node/bucket $p$ at level $(l)$ to its children. This total loss is added to the standard cross-entropy loss for training auto-regressive language models with a regularization hyperparameter $\alpha$. In total, this clustering improves memory utilization by localizing related information within shared memory regions, reducing interference across unrelated tokens, and enabling more stable long-context information retention under a fixed memory budget.

Both the number of buckets and the number of KV slots per bucket remain constant throughout inference, ensuring a bounded memory footprint that does not grow with sequence length. Rather than appending new KV pairs or performing hard eviction, tokens assigned to a bucket are integrated into existing KV slots via learnable update mechanisms described in subsequent sections. This design allows the cache to continuously absorb new information while preserving residual information from previously stored tokens.

\subsection{Differentiable Write Selection}
Unlike implementations that append individual key–value pairs to a dynamically growing structure, \methodname maintains a constant memory budget by selectively updating existing representations within a fixed-size set of memory addresses~\citep{neural-turing-machine,memory-networks, graves2016hybrid}. When updating the cache with a new key-value pair, a specific memory index (we interchangeably use the term \textit{``bucket''} here-forth) is chosen, within which information about the new pair is integrated using the router function. Then the update follows a two-step procedure.

\textbf{Learnable Slot Selection Policy.}
Once an incoming token $\vx_t$ has been routed to a bucket, we select a specific \emph{memory slot} within the assigned bucket for updating. Let
\(
\{\widetilde{\mK}_{t-1,y}, \widetilde{\mV}_{t-1,y}\}_{y=1}^{Y}
\)
denote the $Y$ key-value (KV) slots stored in the selected bucket at decoding step $t-1$. Slot selection is performed by a learnable policy that computes selection logits over these slots conditioned on $\vx_t$.

Specifically, given the key of the incoming token $\vk_t$, we compute slot-selection logits $\vz_t \in \mathbb{R}^{Y}$ using a shared neural scoring function that compares $\vk_t$ with the stored slot keys:
\begin{equation}
\label{eq:gate select}
z_{t,y}
=
\mW_s \bigl[\, \widetilde{\mK}_{t-1,y} \oplus \vk_t \,\bigr] + \vb_s,
\end{equation}
where $\mW_s \in \mathbb{R}^{2d}$ and $\vb_s \in \mathbb{R}$ are learnable parameters shared across slots, and $\oplus$ denotes concatenation.

To enable differentiable learning of discrete slot selection, we apply the Gumbel-Softmax relaxation to the logits $\vz_t$~\citep{gumbel-softmax}. The resulting relaxed selection vector $\boldsymbol{\alpha}_t \in \mathbb{R}^{Y}$ is given by
\begin{equation}
\alpha_{t,y}
=
\frac{
\exp\!\left((z_{t,y} + g_{t,y}) / \tau \right)
}{
\sum_{k=1}^{Y}
\exp\!\left((z_{t,k} + g_{t,k}) / \tau \right)
},
\end{equation}
where $g_{t,y} \sim \operatorname{Gumbel}(0,1)$ and $\tau$ denotes the temperature parameter. During inference, a single slot index
\(
y_t' = \arg\max_{y} \alpha_{t,y}
\)
is selected for updating, with gradients are propagated through $\boldsymbol{\alpha}_t$ during training.

\textbf{Sigmoid-Gated Update.}
Once the slot-selection distribution $\vy_t$ is computed and a slot index $y_t'$ is sampled from $\vy_t$, we update the memory content of the selected slot, $\widetilde{\mK}_{t,y_t'}$ and $\widetilde{\mV}_{t,y_t'} \in \mathbb{R}^d$. Rather than directly overwriting stored contents, as is common in cache eviction policies\citep{h2o,snapkv}, we employ a sigmoid-gated update that softly blends incoming token key and value, $\vk_t$ and $\vv_t$, with existing slot representations.

The update gate output $\bm{\gamma}_t \in \mathbb{R}^{d}$ is computed as
\begin{equation}
    \bm{\gamma}_t
    =
    \sigma\!\left(
    \mW_g \bigl[\widetilde{\mK}_{t-1,y_t'} \oplus \vk_t\bigr]
    + \vb_g
    \right),
\end{equation}
where $\mW_g \in \mathbb{R}^{2d}$ and $\vb_g \in \mathbb{R}$ are learnable parameters, and $\oplus$ denotes concatenation. The memory update is applied sparsely to the selected slot using the sampled address $y_t'$. For $(\mA, \va) \in \{(\mK, \vk), (\mV, \vv)\}$, we perform
\begin{equation}
\label{eq:gated_update}
\widetilde{\mA}_{t,y_t'}=\widetilde{\mA}_{t-1,y_t'}+\alpha_{t,y_t'}\left(-\bm{\gamma}_t \widetilde{\mA}_{t-1,y_t'}+\bm{\gamma}_t  \va_t\right),
\end{equation}
where $\alpha_{t,y_t'}$ is the selected component of $\bm{\alpha}_t$. This gated formulation enables fine-grained control over memory updates, allowing the model to preserve important information when $\bm{\gamma}_t \approx \mathbf{1}$ and overwrite stale or less relevant content when $\bm{\gamma}_t \approx \mathbf{0}$. Similar gating mechanisms have been shown to stabilize sequential memory updates in recurrent and state-space models~\citep{chung2014empirical, yang2024gated}, and our formulation adapts this principle to slot-based KV-cache management under a fixed memory budget.

\subsection{Parallelization of Sigmoid-Gated Memory Updates}
\label{sec:parallel_gated_update}
During auto-regressive generation, attention and KV-cache updates are inherently sequential, which poses no practical limitation. However, the prefilling stage requires KV-cache updates to be executed in parallel across tokens to achieve efficient inference. In \methodname, memory updates follow a GRU-style gated recurrence~\citep{cho2014learning}:
\begin{equation}
 \widetilde{\mA}_{t,y_t'} 
        = \left(1 - \alpha_{t,y_t'} \gamma_t \right)\cdot \widetilde{\mA}_{t-1,y_t'} 
        + \left(\alpha_{t,y_t'} \gamma_t\right)\cdot \va_t,
        \label{eq:state_dep_gate}
\end{equation}
where $\vx_t$ is the incoming token representation and $\va_t$ denotes the write
candidate within the selected memory slot $\widetilde{\mA}_{t,y_t'}$. Since the gate output $\alpha_t$, $\bm{\gamma}_t$ depends explicitly on the key memory state
$\widetilde{\mK}_{t-1,y_t'}$, the update in~\cref{eq:gate select,eq:state_dep_gate} is \emph{inherently sequential}
in the general case.

In \methodname, the mixture weights $\{w_i^{(y)}\}_{i=1}^L$ are not free
parameters but rather implicitly induced by the sequential gated update
mechanism in~\cref{eq:state_dep_gate}. Specifically, each time a write candidate
$\va_t$ is routed to slot $y$, its contribution to the final slot content
$\widetilde{\mA}_y$ is modulated by two learnt factors: the slot-selection
coefficient $\alpha_{t,y}$, which determines whether the update targets slot $y$,
and the sigmoid gate $\gamma_t$, which controls how strongly the new value is
mixed with the existing memory state. Unrolling the recurrence in
\cref{eq:state_dep_gate} over the $L$ writes routed to slot $y$ yields a convex
combination of past write candidates, where each effective weight
$w_i^{(y)}$ corresponds to the product of the write gate at time $i$ and the
survival of that contribution under subsequent gates. Thus
$\vw^{(y)}$ reflects a data-dependent allocation of memory capacity determined
jointly by routing decisions and gated forgetting, rather than an explicitly
parameterized mixture.

\paragraph{Limits of exact parallelization.} $\gamma_t$ and $ \alpha_t$ depend nonlinearly on $\widetilde{\mK}_{t-1,y_t'}$; the recurrence in~\cref{eq:state_dep_gate} thus admits no associative reformulation that would enable exact parallel evaluation over time. This dependence is intrinsic and can not be removed without approximation~\citep{gu2021efficiently, martin2017parallelizing}. As a result, fully state-dependent gating inherently sacrifices parallelism in exchange for more expressive and adaptive memory control.

\paragraph{Re-parameterized gates.}
To recover parallelism, we decouple gate computation from the recurrent memory
state~\citep{katharopoulos2020transformers} and restrict it to depend only on quantities that are available in
parallel. Concretely, both the slot-selection weights and the update gates are
computed using a re-parameterized formulation:
\begin{equation}
    \begin{split}
        \alpha_{t}
        &=
        \operatorname{GS}\!\left(
        \mW_s \bigl[\, \widetilde{\mK}_{t_0,y} \oplus \vk_t \,\bigr] + \vb_s
        \right), \\
        \gamma_t
        &=
        \sigma\!\left(
        \mW_g \bigl[\, \widetilde{\mK}_{t_0,y} \oplus \vk_t \,\bigr]
        + \vb_g
        \right),
    \end{split}
    \label{eq:input_only_gate}
\end{equation}
where $\operatorname{GS}(\cdot)$ denotes the Gumbel-Softmax operator and $t_0$
corresponds to the most recent synchronization point of the KV cache. Under this
formulation, gate values are independent of intermediate updates
within the current sequence.

We first group together all tokens $\vx_t$ that are routed to the same leaf bucket.
Within each bucket, memory updates for different slots are independent. For a
given slot $y$, let $\{t_1, \dots, t_{T_y}\}$ denote the time indices of tokens
assigned to slot $y$. Under~\cref{eq:input_only_gate}, the update for slot $y$
admits the following associative reformulation:
\begin{equation}
\mA_{t} = \left(\prod_{k=1}^{t} \Gamma^{(y)}_k \right)\mA^{(y)}_{0}
+
\sum_{i=1}^{t}
\left(
(1-\Gamma^{(y)}_i)
\prod_{k=i+1}^{t} \Gamma^{(y)}_k
\right)\va^{(y)}_i,
\label{eq:affine_scan}
\end{equation}
where the scalar coefficients $\Gamma^{(y)}_k$ are defined as
\begin{equation}
\Gamma^{(y)}_k =
\begin{cases}
\gamma_k, & \text{if } y_k' = y, \\
1, & \text{otherwise},
\end{cases}
\end{equation}
and $\va^{(y)}_i$ denotes the write candidate at time $t_i$ for slot $y$. This
recurrence corresponds to a prefix-scan~\citep{blelloch1990prefix} over affine
transformations and can be evaluated in $\mathcal{O}(\log T_y)$ depth using
standard parallel scan algorithms.

\subsection{Adaptive Top-$M$ Retrieval via a MDP}
Rather than attending to the entire KV cache, \methodname employs a sparse
retrieval mechanism that selects only a subset of relevant memory buckets for each query. Existing sparse attention methods typically retrieve a fixed number of tokens (e.g., Top-$k$) ~\citep{child2019generating, reformer}, which is suboptimal: simple queries may require only a small context, while complex reasoning tasks benefit from accessing a larger memory footprint. Consequently, the optimal retrieval budget varies across inputs and tasks~\citep{dehghani2018universal}. We therefore formulate the problem of selecting the number of retrieved buckets $B$ as a Markov Decision Process (MDP)~\citep{bellman1957markovian}.

\paragraph{State Formulation.}
The state $\vs_t$ summarizes the current input context using the first $N_s$
tokens of the sequence or a compact representation thereof (e.g., a pooled hidden
state). This representation encodes the contextual complexity that governs the
required retrieval budget.

\paragraph{Policy and Differentiable Transition.}
We implement adaptive retrieval budget selection via a Markovian policy that
iteratively decides whether to increase the beam width (i.e., the number of
retrieved buckets) or to stop. Given the hidden states of the current input
sequence, an initial context representation is constructed by averaging the
first $\ell$ tokens:
\begin{equation}
\vs_0
=
\frac{1}{\ell}\sum_{t=1}^{\ell} \vh_{b,t},
\end{equation}
where $\ell=\min(\texttt{context\_tokens},T)$ and $\vh_{b,t}$ denotes the hidden
state of token $t$. This pooled context is used to
initialize a single-step LSTM state $(\vh_0,\vc_0)$, which serves as a compact
Markov state for subsequent decisions~\citep{hausknecht2015deep}.

Starting from a base width $M_0=1$, the policy rolls forward for a maximum
of $M_{\max}$ steps (bounded by the total number of leaf buckets). At decision
step $m$, we inject a learned step embedding $\ve_m$ into an LSTM cell to produce
an updated hidden state $\vh_m$, and compute a scalar stopping score
\begin{equation}
g_m = \sigma\!\left(\mW_{\mathrm{bw}}\vh_m + \vb_{\mathrm{bw}}\right)\in[0,1],
\end{equation}
where $g_m$ corresponds to the probability of \emph{continuing} to increase the
beam width. During inference, we increment the beam width if $g_m \ge 0.5$ and
terminate otherwise:
\begin{equation}
M_{m+1} =
\begin{cases}
M_m + 1, & \text{if } g_m \ge 0.5,\\
M_m, & \text{otherwise (stop)}.
\end{cases}
\end{equation}
During training, we additionally employ an $\epsilon$-greedy exploration strategy \citep{sutton1998reinforcement} with exploration rate $\epsilon=0.1$ that randomly chooses between incrementing and stopping with a small probability, improving the robustness of the learned policy.

In the training stage, to enable end-to-end optimization with the language modeling loss, we retain two
retrieval candidates computed by the router: $\operatorname{Top}(\mathcal{M},M)$
and $\operatorname{Top}(\mathcal{M},M{+}1)$, where $\operatorname{Top}(\mathcal{M}, M)$ denotes the sparse attention outputs
constructed by retrieving the $M$ highest-scoring memory buckets from $\mathcal{M}$
according to the router’s relevance scores, and attending only to the KV pairs
contained within those buckets (i.e., Top-$M$ sparse attention). We then form a differentiable
``soft-cascade'' interpolation using the final decision score $g_m$:
\begin{equation}
\mathcal{C}_{\mathrm{final}}
=
g_m \cdot \operatorname{Top}(\mathcal{M}, M{+}1)
+
(1-g_m)\cdot \operatorname{Top}(\mathcal{M}, M),
\end{equation}
which allows gradients to shape the stopping behavior while approximating the
discrete choice of retrieval budget.

\section{Gated Writes to a Shared Memory Address}
\label{sec:gated_shared_memory}
\methodname writes multiple past values into a fixed set of memory slots.
When the sequence length exceeds the available number of slots, collisions are
unavoidable and multiple write candidates must be compressed into the same
location. Classical eviction policies (e.g., FIFO or LRU) resolve collisions by
discarding values~\citep{snapkv,adakv,shutova2025cache}, implicitly assuming certainty about which information will be
useful in the future. In contrast, a learned gated write resolves collisions by
\emph{softly combining} values~\citep{graves2016hybrid}, allowing the memory content to hedge against
uncertainty about future usage. We analyze memory writes as a decision under uncertainty problem and show
gated linear combinations are Bayes-optimal when multiple values must share a fixed-capacity memory slot.

\subsection{Writing Multiple Values to One Slot}
Consider a memory slot indexed by $y$, whose content
$\widetilde{\mA}_y \in \mathbb{R}^d$ has fixed capacity. Over time, a collection
of $L$ write candidates $\{\va_1,\dots,\va_L\} \subset \mathbb{R}^d$ are routed to this slot and must be stored jointly. Rather than evicting earlier values, we
represent the slot content as a gated linear combination:
\begin{equation}
\widetilde{\mA}_y
=
\sum_{i=1}^{L} w^{(y)}_i \va_i,
\qquad
w^{(y)}_i \ge 0,\quad \sum_{i=1}^{L} w^{(y)}_i = 1,
\label{eq:gated_write}
\end{equation}
where $\vw^{(y)}$ allocates limited capacity across colliding write candidates.
Eviction-based policies correspond to restricting $\vw^{(y)}$ to simplex vertices
(i.e., selecting a single $\va_i$ and discarding the rest).

These weights $w_i^{(y)}$ are induced implicitly by the gated updates in
\cref{eq:state_dep_gate}. If slot $y$ receives write candidates
$\{\va_1,\dots,\va_L\}$ in order, the contribution of $\va_i$ to the
final memory content is
\(
w_i^{(y)}
=
\alpha_i \gamma_i
\prod_{j=i+1}^{L}\bigl(1-\alpha_j \gamma_j\bigr),
\)
where subsequent gated updates attenuate earlier writes.

\paragraph{Usage intensities (unnormalized).}
At the time of writing, the memory does not know which of the colliding write
candidates will be required by downstream computation. We therefore model future
usage via nonnegative \emph{usage intensities}
$u^{(y)}_i \ge 0$, which may represent empirical access counts, accumulated
attention mass, or abstract rates of anticipated relevance. These intensities
encode relative importance but need not sum to one. Normalizing them yields a
distribution
\begin{equation}
\pi^{(y)}_i
=
\frac{u^{(y)}_i}{\sum_{k=1}^{L} u^{(y)}_k},
\qquad
\sum_{i=1}^{L} \pi^{(y)}_i = 1,
\label{eq:pi_from_u}
\end{equation}
which induces a random index $r \sim \vpi^{(y)}$ corresponding to the write candidate queried in the future.

\paragraph{Objective.}
Because a slot’s content must be written before the model knows which of the
colliding candidates will be accessed later, we view memory writing as a decision
under uncertain future access. In downstream computation, the stored vector
$\widetilde{\mA}_y$ is accessed through attention and acts as a proxy for one of
the routed write candidates $\va_1,\dots,\va_L$. We therefore choose
$\widetilde{\mA}_y$ to minimize the expected reconstruction error of the
eventually queried candidate. Specifically, we define the \emph{expected
reconstruction distortion} as
\begin{equation}
D_L^{(y)}
\;\triangleq\;
\min_{\vw^{(y)} \in \Delta^{L-1}}
\;
\mathbb{E}_{r \sim \vpi^{(y)}}
\left[
\left\|
\va_r - \widetilde{\mA}_y
\right\|_2^2
\right],
\label{eq:objective_shared}
\end{equation}
where \(\vpi^{(y)}\) encodes uncertainty over future access patterns induced by
attention. Under squared error loss, this objective corresponds to the
Bayes-optimal memory representation for an unknown future query~\citep{bishop2006pattern}. In contrast, eviction-based policies implicitly assume certainty about which value will be needed, discarding all others, which can be suboptimal when access is task-dependent.

\begin{table*}[ht!]
    \caption{Results on language modeling and zero-shot common-sense reasoning tasks.}
    \vspace{-0.5\baselineskip}
    \label{tab:harness}
    \centering
    \resizebox{\linewidth}{!}{
    
    \begin{tabular}{cl|cc|cccccccc|c}
        \toprule
        & \multirow{2}{*}{\textbf{Method}} & \textbf{Wiki.}  &  \textbf{LMB.} & \textbf{LMB.} & \textbf{PIQA} &    \textbf{Hella.} & \textbf{Wino.} & \textbf{ARC-c} &  \textbf{ARC-e} & \textbf{SIQA} & \textbf{BoolQ} & \multirow{2}{*}{\textbf{Avg.}} \\
        & & $\mathsf{ppl}\downarrow$ & $\mathsf{ppl}\downarrow$ &  $\mathsf{acc}\uparrow$ & $\mathsf{acc}\uparrow$ & $\mathsf{acc\_n}\uparrow$ & $\mathsf{acc}\uparrow$ & $\mathsf{acc}\uparrow$ & $\mathsf{acc\_n}\uparrow$ &  $\mathsf{acc}\uparrow$ & $\mathsf{acc}\uparrow$ & \\
        \midrule
        \midrule
        \multirow{6}{*}{\rotatebox{90}{\texttt{Llama-3.1-8B}}} & Full Attention & \textbf{7.54} & \textbf{3.14} & 74.54 & \textbf{81.12} & 79.29 & \textbf{74.19} & 55.12 & 82.53 & \textbf{48.21} & \textbf{83.15} & \textbf{72.27} \\
        & SWA (256) & \textbf{7.54} & 3.15 & \textbf{74.81} & 80.52 & 79.34 & 73.88 & 54.95 & 82.41 & 48.06 & 83.09 & 72.13 \\
        & StreamingLLM (4, 256) & \textbf{7.54} & 3.15 & \textbf{74.81} & 80.52 & 79.34 & 73.88 & 54.95 & 82.41 & 48.06 & 83.09 & 72.13 \\
        & Quantization & \textbf{7.54} & \textbf{3.14} & 74.54 & \textbf{81.12} & 79.29 & \textbf{74.19} & 55.12 & 82.53 & \textbf{48.21} & \textbf{83.15} & \textbf{72.27} \\
        & Offloading & 7.54 & 3.15 & \textbf{74.81} & 80.52 & 79.34 & 73.88 & 54.95 & 82.41 & 48.06 & 83.09 & 72.13 \\
        & \methodname~\textbf{(Ours)} & {7.86} & {3.21} & 74.36 & 80.89 & \textbf{79.57} & 73.86 & \textbf{55.42} & \textbf{82.55} & 47.45 & 83.01 & 71.89 \\
        \bottomrule
    \end{tabular}
    }
    \vspace{-\baselineskip}
\end{table*}

\begin{table}[ht!]
    \caption{Results on shorter-context recall-intensive tasks.}
    \vspace{-0.5\baselineskip}
    \label{tab:recall}
    \centering
    \resizebox{\linewidth}{!}{
    \begin{tabular}{cl|cccccc|c}
        \toprule
        & {\textbf{Method}} & \textbf{FDA} & \textbf{SWDE} & \textbf{SQuAD} & \textbf{TQA} & \textbf{NQ} & \textbf{Drop} & {\textbf{Avg.}} \\
        \midrule
        \midrule
        
        \multirow{6}{*}{\rotatebox{90}{\texttt{Llama-3.1-8B}}} & Full Attention & \textbf{22.07} & 31.62 & 50.07 & \textbf{32.93} & 10.86 & \textbf{4.29} & 25.31 \\ 
        & SWA (256) & 7.08 & 20.11 & 50.09 & 20.07 & 5.73 & 1.48 & 17.41 \\
        & StreamingLLM (4, 256) & 22.06 & 31.63 & 50.11 & 32.91 & 10.82 & 4.30 & 25.31 \\
        & Quantization & 21.97 & 31.47 & \textbf{50.12} & 32.88 & 10.92 & 4.27 & 25.29 \\
        & Offloading & 21.88 & 31.42 & 50.11 & 32.83 & 10.91 & 4.25 & 25.23 \\ 
        & \methodname~\textbf{(Ours)} & 21.78 & \textbf{33.21} & 50.07 & 32.18 & \textbf{10.94} & 4.23 & \textbf{25.40} \\
        \bottomrule
    \end{tabular}
    }
    \vspace{-\baselineskip}
\end{table}

\begin{table*}[ht!]
    \caption{Results on \textsc{LongBench}~\citep{longbench}.}
    \vspace{-0.5\baselineskip}
    \label{tab:longbench}
    \centering
    \resizebox{\linewidth}{!}{
    \begin{tabular}{cl|ccc|ccc|ccc|ccc|cc|c}
    \toprule
    & \multirow{2}{*}{\textbf{Method}} 
    & \multicolumn{3}{c|}{\textbf{Single-Doc QA}} 
    & \multicolumn{3}{c|}{\textbf{Multi-Doc QA}} 
    & \multicolumn{3}{c|}{\textbf{Summarization}} 
    & \multicolumn{3}{c|}{\textbf{Few-shot}} 
    & \multicolumn{2}{c|}{\textbf{Code}} 
    & \multirow{2}{*}{\textbf{Avg.}} \\
    & & NQA & QQA & MFQ 
    & HQA & 2WM & Mus 
    & GvR & QMS & MNs
    & TRC & TQA & SSM 
    & LCC & RBP & \\ 
    \midrule
    \midrule
    \multirow{6}{*}{\rotatebox{90}{\texttt{Llama-3.1-8B}}} & Full Attention & 2.43 & 4.86 & 11.25 & \textbf{6.35} & 8.41 & 3.93 & 12.90 & 18.01 & 14.09 & 38.00 & \textbf{44.25} & \textbf{32.23} & \textbf{20.04} & \textbf{22.45} & 17.09 \\
    & SWA (256) & 0.20 & 0.31 & 0.80 & 1.15 & 1.36 & 0.77 & 0.68 & 3.21 & 0.07 & 0.00 & 13.41 & 6.52 & 6.32 & 5.46 & 2.88\\
    & StreamingLLM (4, 256) & 0.90 & 1.65 & 9.12 & 4.02 & 5.56 & 0.79 & 5.12 & 15.28 & 0.27 & 29.50 & 38.60 & 17.95 & 15.96 & 6.91 & 10.76\\
    & Quantization & 1.93 & 5.14 & 11.29 & 6.21 & 8.48 & \textbf{4.01} & 12.25 & 15.07 & \textbf{14.10} & 38.10 & 44.11 & 32.13 & 19.96 & 20.98 & 16.98 \\
    & Offloading & 2.38 & 1.51 & 10.85 & 6.07 & 6.83 & 3.88 & 12.43 & 16.54 & \textbf{14.10} & 37.98 & 44.22 & 32.21 & 19.86 & 19.45 & 16.68 \\
    & \methodname~\textbf{(Ours)} & \textbf{2.45} & \textbf{7.11} & \textbf{15.41} & 6.27 & \textbf{9.39} & 3.84 & \textbf{17.65} & \textbf{20.36} & 10.84 & \textbf{62.50} & 43.42 & 28.20 & 11.21 & 13.68 & \textbf{18.02} \\
    \bottomrule
    \end{tabular}
    }
    \vspace{-\baselineskip}
\end{table*}

\begin{table}[ht!]
    \caption{\textsc{RULER} results at varying context lengths.}
    \vspace{-0.5\baselineskip}
    \label{tab:ruler}
    \centering
    \resizebox{\linewidth}{!}{
    \begin{tabular}{cl|cccccc|c}
    \toprule
    & \textbf{Method}
    & 4\textsf{K} & 8\textsf{K} & 16\textsf{K} & 32\textsf{K} & 64\textsf{K} & 128\textsf{K} & \textbf{Avg.} \\
    \midrule
    \midrule
    \multirow{6}{*}{\rotatebox{90}{\texttt{Llama-3.1-8B}}} & Full Attention & 14.23 & 6.90 & 7.85 & 5.04 & 5.84 & OOM & 8.01 \\
    & SWA (256) & 3.47 & 2.01 & 0.01 & 0.01 & 0.00 & 0.00 & 1.10 \\
    & StreamingLLM (4, 256) & 14.55 & 6.86 & 3.01 & 0.00 & 0.00 & 0.00 & 4.88\\
    & Quantization & 14.23 & 6.91 & 7.83 & 5.04 & 5.85 & OOM  & 8.01 \\ 
    & Offloading & 14.22 & 6.88 & 7.86 & 5.02 & 5.83 & OOM  & 8.00 \\ 
    & \methodname~\textbf{(Ours)} & \textbf{32.67} & \textbf{17.70} & \textbf{10.01} & \textbf{8.55} & \textbf{6.65} & \textbf{5.78} & \textbf{15.12} \\
    \bottomrule
    \end{tabular}
    }
    \vspace{-\baselineskip}
\end{table}

\subsection{Learning Optimal Gated Weights}

For analytical tractability, we adopt a de-correlated model:
\begin{equation}
\mathbb{E}[\va_i] = 0,
\qquad
\mathbb{E}[\va_i^\top \va_j] =
\begin{cases}
\mSigma, & i=j,\\
0, & i\neq j.
\end{cases}
\label{eq:uncorr_shared}
\end{equation}

The expected reconstruction error simplifies to
\begin{equation}
\begin{split}
&\underset{r \sim \vpi^{(y)}}{\mathbb{E}}\!
\left[
\left\|
\va_r - \widetilde{\mA}_y
\right\|_2^2
\right]\\
=&\mathrm{tr}(\mSigma)
\left(
1 + \|\vw^{(y)}\|_2^2 - 2 \langle \vpi^{(y)}, \vw^{(y)} \rangle
\right).
\end{split}
\label{eq:avg_error_shared}
\end{equation}

This convex quadratic program admits a unique minimizer
\begin{equation}
\vw^{(y)\star} = \vpi^{(y)}
\quad\Longleftrightarrow\quad
w^{(y)\star}_i
=
\frac{u^{(y)}_i}{\sum_{k=1}^{L} u^{(y)}_k}.
\label{eq:optimal_weights_shared}
\end{equation}
which allocates memory capacity proportionally to the anticipated usage
intensity of each write candidate, and yields the minimum achievable
distortion when \(L\) values must be jointly stored in a single memory slot.

\subsection{Information-Theoretic View}
Assume additionally that the write candidates are routed to the same memory address
are i.i.d.\ Gaussian,
\(
\va_j \overset{\text{i.i.d.}}{\sim} \mathcal{N}(\mathbf{0}, \mSigma).
\)
For Gaussian variables, minimizing mean-squared reconstruction error is
equivalent to minimizing conditional entropy~\citep{cover1999elements}. 
Using the uncorrelated assumption in~\cref{eq:uncorr_shared} and the optimal
gated weights \(\vw^\star = \vpi\), the residual covariance scales with
\(1 - \|\vpi\|_2^2\). This yields the conditional mutual information
\begin{equation}
I(\va_r; \mA_a \mid r)
\ge
-\frac{d}{2}
\log\!\left(1 - \|\vpi\|_2^2\right).
\label{eq:mi_shared}
\end{equation}
where \(r \sim \vpi\) denotes the index of the future queried value.
Thus, when future usage concentrates on a small subset of values
(i.e., large \(\|\vpi\|_2^2\)), a single memory address preserves more
information about the to-be-accessed content.

\subsection{Why Learned Gated Writes Beat Eviction}
Eviction-based policies~\citep{snapkv,adakv,shutova2025cache} implicitly set $\vw^{(y)}$ to a one-hot vector, discarding
all but one write candidate and assuming certainty about future access. In
contrast, \cref{eq:optimal_weights_shared} shows that under stochastic or
task-dependent usage, the optimal strategy is to \emph{share} memory capacity in
proportion to anticipated utility. Learned gating provides a mechanism to
estimate usage intensities from data and implement this optimal allocation,
whereas fixed eviction rules can not adapt to semantic or task-specific notions of
importance.

Beyond optimality under uncertainty, gated writes also yield more realistic temporal dynamics. In practice, usage intensity evolves over time, with older information typically becoming less relevant as new context arrives. A learned sigmoid gate naturally captures this behavior by smoothly decaying the contribution of past memory content during each write, rather than discarding values abruptly. Our analysis isolates the write operation and does not claim optimality of the full routing or attention mechanism; nonetheless, under minimal uncertainty assumptions about future access, gated writes form a strictly more expressive and principled policy class than eviction-based updates.

\section{Experimental Results}

\subsection{Setup}

\paragraph{Benchmarks.} To evaluate \methodname, we conduct evaluations on a number of different benchmarks, such as a number of tasks using \texttt{lm-eval}~\citep{eval-harness}, the \textsc{LongBench}~\citep{longbench} benchmark suite, and the \textsc{RULER} suite of tasks~\citep{ruler} up to a context length of 128\textsf{K}. 

\paragraph{Baselines.} We compare our method against a number of standard baselines, such as full attention implemented with \textsc{FlashAttention}~\citep{dao2022flashattention, dao2024flashattention2, dao2024flashattention3}, sliding window and sink caches/StreamingLLM~\citep{attention-sinks}, a quantized caching method in \textsc{KIVI}~\citep{kivi}, as well as a cache offloading to CPU memory approach where only the current layer's cache is stored on the GPU. For sliding window and sink caches, we set the total number of tokens to be maintained in the cache to 256; accordingly, \methodname is also instantiated with 256 memory slots. This is constructed as a 4-level tree structure with 4 children per node, leading to 256 leaf buckets.

\paragraph{Model.} We conduct our experiments on \texttt{Llama3}~\citep{llama3}. Due to resource constraints, we focus on the use of the 8B-sized model. 
For reasons of training the Gumbel-Write and MDP-Read modules not present in the original model, we pre-train models on a 10B token subset of the \texttt{FineWeb-Edu} dataset~\citep{fineweb}.

\subsection{Performance and Results}

\cref{tab:harness} first presents performance on non-generative tasks and thus no direct usage of a cache for full attention. In this case, performance should not deviate from the full-attention baseline. This is clearly represented with the minimal changes in average performance and perplexity of standard caching techniques; similarly, results from \methodname shows that directly adapting a pre-trained model to utilize our structure does not experience a noticeable change either.

On short-context retrieval tasks (\cref{tab:recall}) where the context length generally will not constitute a memory bottleneck, using caching techniques can lead to drops in performance on some tasks, such as naive sliding window caching and offloading to CPU. Meanwhile, quantization and sinks are capable of maintaining performance on these shorter-context settings. \methodname, with no trade-off in performance due to the information mixing technique but gains in memory and latency are observed relative to full KV-cache maintenance.

On real-world, long-context tasks (\cref{tab:longbench}), only quantizing the cache remains comparable to full-attention. \methodname meanwhile sees relative improvements in performance alongside gains in memory and speed, highlighting the major benefits observed through this method.

Finally, \cref{tab:ruler} shows that \methodname can shine by not only maintaining performance across lengths where a full KV-cache is used, as others either fail to succeed or run out of memory, but it remains the only method to have any success at lengths where memory becomes an issue. We present an ablation study in \cref{sec:ablation_sparsity_gating} on shorter-context, recall-intensive tasks, analyzing the impact of gated writes and sparsity control mechanisms.

\begin{figure}[ht!]
    \centering
    \includegraphics[width=\linewidth]{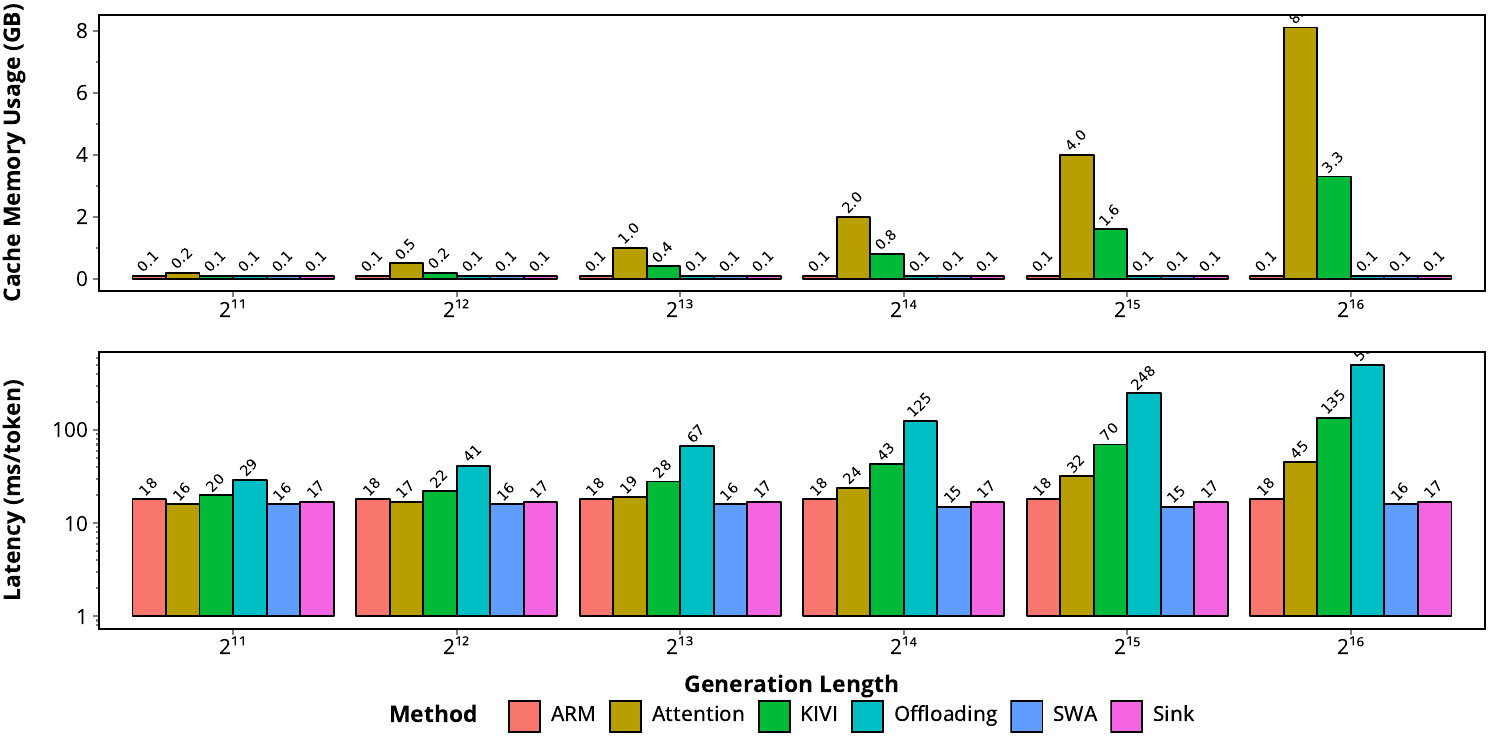}
    \includegraphics[width=\linewidth]{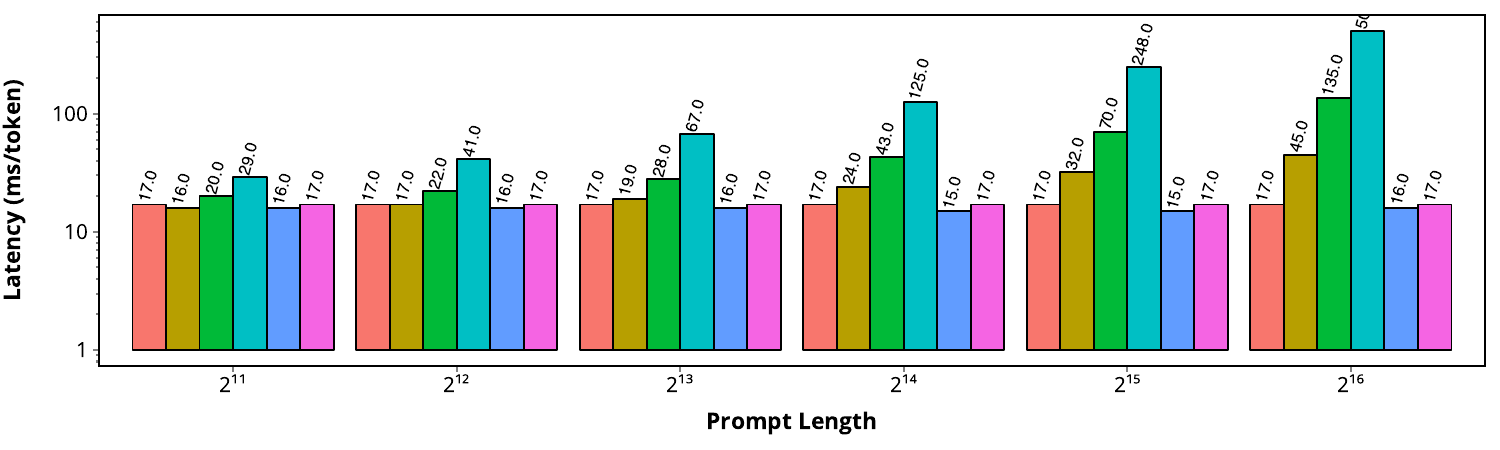}
    \caption{Comparison of memory usage of the KV cache (top) and decoding latency (middle) when generating at various lengths of outputs. Bottom plot presents the average generation latency using fixed-length inputs/prompts.
    Results presented at on log scale.}
    \vspace{-1.5\baselineskip}
    \label{fig:benchmarking}
\end{figure}

\subsection{Benchmarking}

To further demonstrate the benefits of our method, we benchmark \methodname during generation by measuring generation latency across varying prompt lengths as well as maximum memory usage of the KV-cache\footnote{We compute memory consumption of the KV-cache separate from the attention computation.}. As we observe in \cref{fig:benchmarking}, \methodname delivers a significant speedup compared to methods that maintain a full KV-cache (full attention, quantized caching) while on-par with sliding window attention and sink caches. While cache offloading maintains the same memory utilization, it sees significant latency issues relative to other methods (over 10$\times$ higher latency), highlighting the impracticality of such methods for long-context generation. 
This extends to generation when processing longer prompts, where the relative speed of generating each new token does not grow and is therefore comparable with sliding window and sink caches.
Combined with the improved performance relative to specific cache eviction policies, these results further reinforce the effectiveness of our approach.

\section{Conclusion}

We presented \methodname, a method to make tree-structured KV caches fully differentiable and adaptive. By leveraging Gumbel-Softmax for write operations and a differentiable MDP for read operations, we enable LLMs to learn optimal memory management strategies end-to-end. Experimental results on a variety of long-context and commonsense reasoning tasks with a \texttt{Llama-3} backbone demonstrate both the effectiveness of the approach through performance improvements relative to standard eviction baselines as well as the scalability compared to quantization and offloading approaches. Latency and memory usage benchmarking further lends credit to this claim. With these promising results, future work can explore infinite-context settings or possibly sub-linear memory growth.

\section*{Impact Statement}

This paper introduces a new KV caching structure for more effective inference with large language models. While there may be some downstream applications that can merit some greater investigation in terms of downstream usage (ex. responsible and fair usage of LLMs), these are inherent considerations that do not directly stem from the novelties introduced within this work and therefore we do not believe necessitate specific mention here.

\section*{Acknowledgements}
Q. Zeng, R. Fang, G. Xu, Z. Jing and B. Wang are supported by the Natural Sciences and Engineering Research Council of Canada (NSERC) Discovery Grants program. J. Huang is supported by the NSERC Canada Graduate Scholarships program (reference numbers 589326 and 611521-2025). The authors also would like to acknowledge Yu Chen (NTU) for helpful discussions on designing the learnable sparsity approach and Yanan Wang for technical help.

\nocite{fang2026saga, DBLP:conf/iclr/FangL0ZDP0025, DBLP:conf/aaai/FangWPZZWCMTLW26, DBLP:conf/aaai/FangWPZZWCMTLW26, li2026when, li2026versatile, cai2026fuse, DBLP:conf/aaai/PuXFB0025}
\nocite{huang-etal-2025-robot, lu2025mamba, lu2025mambamodulationlengthgeneralization, huang-2025-well, huang2025calibrated, huang-etal-2026-investigating, zeng2025zetaleveragingzordercurves}

\bibliography{refs}
\bibliographystyle{icml2026}

\clearpage
\onecolumn

\appendix

\section{Information-Theoretic View}
Assume additionally that the write candidates routed to the same memory address
are i.i.d.\ Gaussian,
\(
\va_j \overset{\text{i.i.d.}}{\sim} \mathcal{N}(\mathbf{0}, \mSigma).
\)
For Gaussian variables, minimizing mean-squared reconstruction error is
equivalent to minimizing conditional entropy~\citep{cover1999elements}.
To avoid mixture effects induced by the random query index, we consider the
conditional mutual information given the future access index \(r\):
\begin{equation}
I(\va_r;\widetilde{\mA}_y \mid r)
\;\triangleq\;
\mathbb{E}_{k \sim \vpi}
\!\left[
I(\va_k;\widetilde{\mA}_y)
\right].
\end{equation}
Using the uncorrelated assumption in~\cref{eq:uncorr_shared} and the optimal
gated weights \(\vw^\star = \vpi\), the residual covariance conditioned on \(r=k\)
scales with \(1 - \|\vpi\|_2^2\), yielding
\begin{equation}
I(\va_r;\widetilde{\mA}_y \mid r)
\ge
-\frac{d}{2}
\log\!\left(1 - \|\vpi\|_2^2\right).
\end{equation}
Thus, when future usage concentrates on a small subset of values
(i.e., large \(\|\vpi\|_2^2\)), a single shared memory slot preserves more
information about the value that will be queried later.

\subsection{Proof of Information-Theoretic View}

Assume \(\va_i \overset{\text{i.i.d.}}{\sim}\mathcal{N}(\mathbf{0},\mSigma)\) and the de-correlated model in~\cref{eq:uncorr_shared}. Let the stored proxy be
\(
\widetilde{\mA}_y=\sum_{i=1}^L w_i \va_i
\)
with \(\vw\in\Delta^{L-1}\), and let \(r\sim\vpi\) denote the future queried index.
Define the residual \(\ve \triangleq \va_r-\widetilde{\mA}_y\).

\paragraph{Residual covariance conditioned on the query.}
Conditioned on \(r=k\),
\[
\ve
=\va_k-\sum_{i=1}^L w_i\va_i
=(1-w_k)\va_k-\sum_{i\neq k} w_i\va_i.
\]
By independence and~\cref{eq:uncorr_shared},
\[
\mathrm{Cov}(\ve\mid r=k)
=\Bigl((1-w_k)^2+\sum_{i\neq k} w_i^2\Bigr)\mSigma
=\Bigl(1-2w_k+\|\vw\|_2^2\Bigr)\mSigma.
\]

\paragraph{Mixture-averaged residual covariance.}
Averaging over \(r\sim\vpi\) gives
\begin{equation}
\mathrm{Cov}(\ve)
=
\mathbb{E}_{r}\!\left[\mathrm{Cov}(\ve\mid r)\right]
=
\Bigl(1+\|\vw\|_2^2-2\langle\vpi,\vw\rangle\Bigr)\mSigma,
\label{eq:res_cov_app}
\end{equation}
since \(\mathbb{E}_{r\sim\vpi}[w_r]=\langle\vpi,\vw\rangle\).

\paragraph{Conditional entropy.}
Since each \(\va_k\) is Gaussian and \(\widetilde{\mA}_y\) is a linear
combination of Gaussians, \(\ve\mid r=k\) is Gaussian with covariance
\(\mathrm{Cov}(\ve\mid r=k)\). Using the Gaussian entropy formula,
\[
H(\ve\mid r=k)
=
\frac{1}{2}\log\det\!\Bigl((2\pi e)\,\mathrm{Cov}(\ve\mid r=k)\Bigr).
\]

\paragraph{Gaussian upper bound on conditional entropy.}
For any random vector with covariance \(\mC\), its differential entropy is
upper bounded by that of a Gaussian with the same covariance. Applying this to
\(\ve\) and using~\cref{eq:res_cov_app} yields
\[
H(\va_r\mid \widetilde{\mA}_y, r)
=
\mathbb{E}_{k\sim\vpi}\!\left[H(\ve\mid r=k)\right]
\;\le\;
\frac{1}{2}\log\det\!\Bigl((2\pi e)\,\mathrm{Cov}(\ve)\Bigr).
\]
Moreover, \(\va_r\mid r\) is Gaussian with covariance \(\mSigma\), hence
\[
H(\va_r\mid r)=\frac{1}{2}\log\det((2\pi e)\mSigma).
\]
Therefore, the conditional mutual information satisfies
\begin{align}
I(\va_r;\widetilde{\mA}_y\mid r)
&=
H(\va_r\mid r)-H(\va_r\mid \widetilde{\mA}_y, r)
\nonumber\\
&\ge
\frac{1}{2}\log\det((2\pi e)\mSigma)
-\frac{1}{2}\log\det\!\Bigl((2\pi e)\,\mathrm{Cov}(\ve)\Bigr)
\nonumber\\
&=
-\frac{d}{2}\log\!\Bigl(1+\|\vw\|_2^2-2\langle\vpi,\vw\rangle\Bigr).
\label{eq:mi_lb_app}
\end{align}

\paragraph{Optimal weights.}
By \cref{eq:optimal_weights_shared}, the unique minimizer of the expected
reconstruction error satisfies \(\vw^\star=\vpi\). Substituting \(\vw=\vpi\) gives
\[
1+\|\vw\|_2^2-2\langle\vpi,\vw\rangle
=
1-\|\vpi\|_2^2.
\]
Plugging into~\cref{eq:mi_lb_app} yields
\[
I(\va_r;\widetilde{\mA}_y\mid r)
\;\ge\;
-\frac{d}{2}
\log\!\left(1-\|\vpi\|_2^2\right),
\]
which matches~\cref{eq:mi_shared} as the stated information-preservation proxy.
$\hfill\Box$

\clearpage
\section{Additional Experimental Details}

Here we list some additional details regarding the different tasks on which we conduct language model evaluation.

\subsection{Experimental Environment Setup}

Training was conducted on two nodes with 4 NVIDIA A100 GPUs with 40GB of memory, connected through NVLink. Inference results are all obtained using a NVIDIA H100 GPU with 80GB of memory. Models are initialized using \texttt{bf16} half-precision format, while quantization uses 4 bits.

\subsection{Additional Training Details}

Training uses sequences with context length of 2048 tokens. We use the AdamW optimizer~\citep{adamw} with a peak learning rate of 4e-4, weight decay of 0.1, and gradient clipping of 1.0. The learning rate follows a cosine annealing schedule with a warm-up period of 1\% of the total steps ($\approx$100M tokens) and a total batch size of 0.5M tokens.

\subsection{Language Model Evaluation Harness Tasks}\label{app:harness_details}

The following are recall-intensive tasks on which we evaluate. All tasks are evaluated directly using accuracy for commonsense reasoning tasks and perplexity for language modeling.

\begin{table*}[ht]
    \centering
    \caption{Harness tasks on which we evaluate.}
    \begin{tabular}{l|cccc}
    \toprule
    \multicolumn{1}{c}{\textbf{Task}} & Task Type \\
    \midrule
     \textsc{PIQA}~\citep{piqa} & Physical Commonsense Reasoning \\
     \textsc{Arc}~\citep{arc} & Commonsense Reasoning \\
     \textsc{HellaSwag}~\citep{hellaswag} & Commonsense Natural Language Inference \\
     \textsc{WinoGrande}~\citep{winogrande} & Pronoun Resolution \\
     \textsc{SIQA}~\citep{siqa} & Social Commonsense Reasoning \\
     \textsc{BoolQ}~\citep{BoolQ} & Yes/No Commonsense QA \\
     \textsc{WikiText}~\citep{wikitext} & Language Modeling \\
     \textsc{LAMBADA}~\citep{lambada} & Text Understanding \\
     \bottomrule
    \end{tabular}
    \label{tab:harness-tasks}
\end{table*}

\subsection{Recall Intensive Tasks}\label{app:recall_details}

The following are recall-intensive tasks on which we evaluate. All tasks are evaluated directly with accuracy reported as the metric of choice.

\begin{table*}[ht]
    \centering
    \caption{Recall-intensive tasks on which we evaluate.}
    \begin{tabular}{l|cccc}
    \toprule
    \multicolumn{1}{c}{\textbf{Task}} & Task Type \\
    \midrule
     \textsc{SWDE}~\citep{swde} & Structure HTML Relation Extraction \\
     \textsc{FDA}~\citep{fda} & PDF Key-Value Retrieval \\
     \textsc{SQuAD}~\citep{squad} & Question Answering \\
     \textsc{TriviaQA}~\citep{triviaqa} & Question Answering \\
     \textsc{Drop}~\citep{drop} & Question Answering \\
     \textsc{Natural Questions}~\citep{naturalquestions} & Question Answering \\
     \bottomrule
    \end{tabular}
    \label{tab:recall-tasks}
\end{table*}

\subsection{LongBench}\label{app:longbench_details}

We evaluate the following tasks from \textsc{LongBench}~\citep{longbench} (\cref{tab:longbench-tasks}).

\begin{table*}[ht]
    \centering
    \caption{Tasks from LongBench on which we evaluate.}
    \resizebox{\linewidth}{!}{
    \begin{tabular}{l|cccc}
    \toprule
    \multicolumn{1}{c}{\textbf{Task}} & Context Type & Average Length & Metric & Data Samples \\
    \midrule
     \textsc{NarrativeQA}~\citep{nqa} & Literature/Film & 18409 & F1 & 200\\
     \textsc{QasperQA}~\citep{qasper} & Science & 3619 & F1 & 200 \\
     \textsc{MultiFieldQA}~\citep{longbench} & Multi-Field & 4559 & F1 & 150 \\
     \textsc{HotpotQA}~\citep{hotpotqa} & Wikipedia & 9151 & F1 & 200 \\
     \textsc{2WikiMultiQA}~\citep{2wikimultiqa} & Wikipedia & 4887 & F1 & 200 \\
     \textsc{Musique}~\citep{musique} & Wikipedia & 11214 & F1 & 200 \\
     \textsc{GovReport}~\citep{govreport} & Government Reports & 8734 & Rouge-L & 200 \\
     \textsc{QMSum}~\citep{qmsum} & Meetings & 10614 & Rouge-L & 200 \\
     \textsc{MultiNews}~\cite{multinews} & News & 2113 & Rouge-L & 200 \\
     \textsc{TRec}~\citep{trec} & Web Questions & 5117 & Accuracy & 200 \\
     \textsc{TriviaQA}~\citep{triviaqa} & Wikipedia/Web & 8209 & F1 & 200 \\
     \textsc{SAMsum}~\citep{samsum} & Dialogue & 6258 & Rouge-L & 200 \\
     \textsc{LCC}~\citep{lcc} & Github & 1235 & Edit Similarity & 500 \\
     \textsc{RepoBench-P}~\citep{repobench} & Github Repositories & 4206 & Edit Similarity & 500 \\
     \bottomrule
    \end{tabular}
    }
    \label{tab:longbench-tasks}
\end{table*}

\subsection{RULER}

\textsc{RULER} comprises of task spanning across four categories: \emph{retrieval}, \emph{multi-hop tracing}, \emph{aggregation}, and \emph{question answering}. We use a publicly available repository\footnote{\url{https://github.com/NVIDIA/RULER}} to generate evaluation examples based on specific input configurations (see \autoref{tab:example-task} for example configurations) that define the length and complexity of each input. In \textsc{RULER}, the task complexity can be thought of as a function of the number of target output tokens and the signal-to-noise ratio in the context. For our experiments, we use the default set of tasks pre-defined by \citet{ruler}.

\begin{table}[ht!]
\centering
\resizebox{0.8\linewidth}{!}{
\begin{tabular}[t]{@{}llp{0.8\linewidth}@{}}
\toprule
\textbf{Task} & \textbf{Configuration} & \textbf{Example} \\
\midrule
\begin{tabular}[t]{@{}l@{}}Single\\NIAH\\(S-NIAH)\end{tabular} & 
\begin{tabular}[t]{@{}l@{}}type\_key = word\\type\_value = number\\type\_haystack = essay\\size\_haystack $\propto$ context length\end{tabular} & 
\begin{tabular}[t]{@{}p{\linewidth}@{}}
\textcolor{lightgray}{(essays) ......} \\
One of the special magic numbers for \textcolor{violet}{long-context} is: \textcolor{orange}{12345}. \textcolor{lightgray}{......}  \\
What is the special magic number for \textcolor{violet}{long-context} mentioned in the provided text?\\
Answer: \textcolor{orange}{12345}
\end{tabular}
\\
\midrule
\begin{tabular}[t]{@{}l@{}}Multi-keys\\NIAH\\(MK-NIAH)\end{tabular}  &
\begin{tabular}[t]{@{}l@{}}num\_keys = 2\\type\_key = word\\type\_value = number\\type\_haystack = essay\\size\_haystack $\propto$ context length\end{tabular} & 
\begin{tabular}[t]{@{}p{\linewidth}@{}}
\textcolor{lightgray}{(essays) ......} \\ 
One of the special magic numbers for \textcolor{violet}{long-context} is: \textcolor{orange}{12345}. \\  
\textcolor{lightgray}{One of the special magic numbers for large-model is: 54321}. \\ 
\textcolor{lightgray}{......}  \\
What is the special magic number for \textcolor{violet}{long-context} mentioned in the provided text?\\
Answer: \textcolor{orange}{12345}
\end{tabular}
\\
\midrule
\begin{tabular}[t]{@{}l@{}}Multi-values\\NIAH\\(MV-NIAH)\end{tabular} &
\begin{tabular}[t]{@{}l@{}}num\_values = 2\\type\_key = word\\type\_value = number\\type\_haystack = essay\\size\_haystack $\propto$ context length\end{tabular} & 
\begin{tabular}[t]{@{}p{\linewidth}@{}}
\textcolor{lightgray}{(essays) ......} \\ 
One of the special magic numbers for \textcolor{violet}{long-context} is: \textcolor{orange}{12345}. \\  
One of the special magic numbers for \textcolor{violet}{long-context} is: \textcolor{orange}{54321}. \\  
\textcolor{lightgray}{......}  \\
What are all the special magic numbers for \textcolor{violet}{long-context} mentioned in the provided text?\\
Answer: \textcolor{orange}{12345}  \textcolor{orange}{54321}
\end{tabular}
\\
\midrule
\begin{tabular}[t]{@{}l@{}}Multi-queries\\NIAH\\(MQ-NIAH)\end{tabular} &
\begin{tabular}[t]{@{}l@{}}num\_queries = 2\\type\_key = word\\type\_value = number\\type\_haystack = essay\\size\_haystack $\propto$ context length\end{tabular} &  
\begin{tabular}[t]{@{}p{\linewidth}@{}}
\textcolor{lightgray}{(essays) ......} \\ 
One of the special magic numbers for \textcolor{violet}{long-context} is: \textcolor{orange}{12345}. \\  
One of the special magic numbers for \textcolor{violet}{large-model} is: \textcolor{orange}{54321}. \\  
\textcolor{lightgray}{......}  \\
What are all the special magic numbers for \textcolor{violet}{long-context} and \textcolor{violet}{large-model} mentioned in the provided text?\\
Answer: \textcolor{orange}{12345}  \textcolor{orange}{54321}
\end{tabular}
\\
\midrule
\begin{tabular}[t]{@{}l@{}}Variable\\Tracking\\(VT)\end{tabular} &
\begin{tabular}[t]{@{}l@{}}num\_chains = 2\\num\_hops = 2\\size\_noises $\propto$ context length\end{tabular} &
\begin{tabular}[t]{@{}p{\linewidth}@{}}
\textcolor{lightgray}{(noises) ......} \\
VAR \textcolor{orange}{X1} = \textcolor{violet}{12345} \textcolor{lightgray}{...... VAR Y1 = 54321 ......}  \\
VAR \textcolor{orange}{X2} = \textcolor{orange}{X1} \textcolor{lightgray}{...... VAR Y2 = Y1 ......} \\
VAR \textcolor{orange}{X3} = \textcolor{orange}{X2} \textcolor{lightgray}{...... VAR Y3 = Y2 ......} \\
Find all variables that are assigned the value \textcolor{violet}{12345}. \\
Answer: \textcolor{orange}{X1 X2 X3}
\end{tabular}
\\
\midrule
\begin{tabular}[t]{@{}l@{}}Common Words\\Extraction\\(CWE)\end{tabular} &
\begin{tabular}[t]{@{}l@{}}freq\_cw = 2, freq\_ucw = 1\\num\_cw = 10\\num\_ucw $\propto$ context length\end{tabular} & 
\begin{tabular}[t]{@{}p{\linewidth}@{}}
\textcolor{orange}{aaa} \textcolor{lightgray}{bbb} \textcolor{orange}{ccc} \textcolor{orange}{aaa} \textcolor{lightgray}{ddd} \textcolor{lightgray}{eee} \textcolor{orange}{ccc} \textcolor{lightgray}{fff} \textcolor{lightgray}{ggg} 
\textcolor{lightgray}{hhh} \textcolor{orange}{iii} \textcolor{orange}{iii} \textcolor{lightgray}{......}\\
What are the 10 most common words in the above list? \\
Answer: \textcolor{orange}{aaa ccc iii ......}
\end{tabular}
\\
\midrule
\begin{tabular}[t]{@{}l@{}}Frequent Words\\Extraction\\(FWE)\end{tabular} &
\begin{tabular}[t]{@{}l@{}}$\gamma$ = 2\\num\_word $\propto$ context length\end{tabular} & 
\begin{tabular}[t]{@{}p{\linewidth}@{}}
\textcolor{orange}{aaa} \textcolor{lightgray}{bbb} \textcolor{orange}{ccc} \textcolor{orange}{aaa} \textcolor{lightgray}{ddd} \textcolor{lightgray}{eee} \textcolor{orange}{ccc} \textcolor{lightgray}{fff} \textcolor{lightgray}{ggg} \textcolor{orange}{aaa} \textcolor{lightgray}{hhh} \textcolor{orange}{aaa} \textcolor{orange}{ccc} \textcolor{orange}{iii} \textcolor{orange}{iii}  \textcolor{lightgray}{......}\\
What are the 3 most frequently appeared words in the above coded text? \\
Answer: \textcolor{orange}{aaa ccc iii}
\end{tabular}
\\
\midrule
\begin{tabular}[t]{@{}l@{}}Question\\Answering\\(QA)\end{tabular} &
\begin{tabular}[t]{@{}l@{}}dataset = SQuAD\\num\_document $\propto$ context length\end{tabular} & 
\begin{tabular}[t]{@{}p{\linewidth}@{}}
\textcolor{lightgray}{Document 1: ...... aaa ......} \\
\textcolor{violet}{Document 2:} \textcolor{lightgray}{......} \textcolor{orange}{bbb} \textcolor{lightgray}{......} \\
\textcolor{lightgray}{Document 3: ...... ccc ......} \\
Question: \textcolor{violet}{question} \\
Answer: \textcolor{orange}{bbb}
\end{tabular}
\\
\bottomrule
\end{tabular}}
\caption{Task examples with flexible configurations in \textsc{Ruler}. 
Different colors highlight \textcolor{violet}{queries}, \textcolor{violet}{keys}, \textcolor{orange}{values}, and \textcolor{lightgray}{distractors} in each example. Examples are retrieved directly from \citet{ruler}.}
\label{tab:example-task}
\end{table}

\subsection{Experimental Reproducibility}

We use the \texttt{lm-eval} package\footnote{\url{https://github.com/EleutherAI/lm-evaluation-harness}} for evaluating our models. We use \texttt{Transformers} version 4.57.1 and \texttt{PyTorch} 2.7.1 for evaluating baselines.


\clearpage
\section{Additional Experiments}
\subsection{Ablation: Adaptive Retrieval Sparsity and Gated KV Updates}
\label{sec:ablation_sparsity_gating}
\begin{table*}[t]
\caption{Ablation study on shorter-context recall-intensive tasks. We analyze the impact of gated write and sparsity control mechanisms.}
\label{tab:ablation_sparsity_gating}
\vspace{-0.5\baselineskip}
\centering
\scriptsize
\resizebox{\linewidth}{!}{
\begin{tabular}{l|cc|cccccc|c}
\toprule
\multirow{1}{*}{\textbf{Variant}}
& \textbf{Gated Write} 
& \textbf{Learnable Sparsity} 
& \textbf{FDA} 
& \textbf{SWDE} 
& \textbf{SQuAD} 
& \textbf{TQA} 
& \textbf{NQ} 
& \textbf{Drop} 
& \textbf{Avg.} \\
\midrule
FIFO (256)
 & \texttimes & \texttimes 
 & 7.08 & 20.11 & 50.09 & 20.07 & 5.73 & 1.48 & 17.41 \\

Fixed Sparsity ($1/2$)
 & \checkmark & \texttimes 
 & 21.13 & 32.65 & 50.07 & 29.95 & 8.71 & 4.02 & 24.42 \\

Fixed Sparsity ($1/4$)
 & \checkmark & \texttimes 
 & 20.15 & 31.89 & 50.08 & 29.46 & 7.78 & 4.27 & 24.94 \\

\methodname\ (Learnable Sparsity)
 & \checkmark & \checkmark 
 & \textbf{21.78} & \textbf{33.21} & 50.07 & \textbf{32.18} & \textbf{10.94} & \textbf{4.23} & \textbf{25.40} \\
\bottomrule
\end{tabular}
}
\vspace{-\baselineskip}
\end{table*}

Table~\ref{tab:recall} studies the impact of retrieval sparsity and KV-cache update
policies on recall-intensive benchmarks. We ablate two orthogonal design choices:
(i) replacing the learned, adaptive retrieval budget with fixed sparsity ratios,
and (ii) replacing gated KV updates with a FIFO eviction policy.

Comparing FIFO (256) with fixed-sparsity variants highlights the importance of the
write mechanism. While all methods operate under comparable memory budgets, FIFO
eviction leads to severe performance degradation across all tasks, indicating
that hard replacement of KV entries discards recall-critical information.
Introducing gated KV updates yields large gains even with fixed sparsity, showing
that softly combining memory content is essential for preserving useful signals.

Next, comparing fixed sparsity ratios ($1/2$ and $1/4$) with \methodname
demonstrates the benefit of adaptive retrieval. Fixed ratios impose a uniform
retrieval budget across all inputs, which is suboptimal given varying contextual
complexity. In contrast, the MDP-based policy in \methodname dynamically adjusts
the number of retrieved buckets on a per-input basis, achieving the best average
performance across datasets.

Empirically, we observe that the learned sparsity is not concentrated around a single
fixed ratio: some inputs trigger retrieval budgets larger than $1/2$, while others
require substantially fewer than $1/4$ of the buckets. This variability reflects
differences in information density and reasoning demands across inputs, confirming
that no single fixed sparsity level is universally optimal. Overall, these results
support adaptive, input-dependent sparsity combined with gated KV updates as a
principled alternative to both static sparse attention and eviction-based cache
management.

\end{document}

%% file: math_commands.tex
\usepackage{amsmath,amsfonts,bm}

\def\eqref#1{equation~\ref{#1}}

\def\1{\bm{1}}

\def\va{{\bm{a}}}
\def\vb{{\bm{b}}}
\def\vc{{\bm{c}}}

\def\ve{{\bm{e}}}

\def\vh{{\bm{h}}}

\def\vk{{\bm{k}}}

\def\vo{{\bm{o}}}
\def\vp{{\bm{p}}}
\def\vq{{\bm{q}}}

\def\vs{{\bm{s}}}

\def\vv{{\bm{v}}}
\def\vw{{\bm{w}}}
\def\vx{{\bm{x}}}
\def\vy{{\bm{y}}}
\def\vz{{\bm{z}}}

\def\mA{{\bm{A}}}

\def\mC{{\bm{C}}}

\def\mK{{\bm{K}}}

\def\mQ{{\bm{Q}}}

\def\mV{{\bm{V}}}
\def\mW{{\bm{W}}}
\def\mX{{\bm{X}}}

\def\mSigma{{\bm{\Sigma}}}

\DeclareMathAlphabet{\mathsfit}{\encodingdefault}{\sfdefault}{m}{sl}
\SetMathAlphabet{\mathsfit}{bold}{\encodingdefault}{\sfdefault}{bx}{n}

\def\vpi{\bm{\pi}} 

\def\mSigma{\bm{\Sigma}} 